\documentclass[letterpaper]{article} 
\usepackage{aaai23}  
\usepackage{times}  
\usepackage{helvet}  
\usepackage{courier}  
\usepackage[hyphens]{url}  
\usepackage{graphicx} 
\usepackage{natbib}  
\usepackage{caption} 
\usepackage{algorithm}
\usepackage{algorithmic}

\usepackage{newfloat}
\usepackage{listings}
\DeclareCaptionStyle{ruled}{labelfont=normalfont,labelsep=colon,strut=off} 
\floatstyle{ruled}
\newfloat{listing}{tb}{lst}{}
\floatname{listing}{Listing}
\usepackage{booktabs}
\usepackage{amsmath}
\usepackage{multirow}
\usepackage{amssymb}
\usepackage{bm}
\usepackage{cleveref}
\usepackage{bbold}

\title{Learning to Defer with Limited Expert Predictions}
\author {
    Patrick Hemmer\equalcontrib,
    Lukas Thede\equalcontrib,
    Michael Vössing,
    Johannes Jakubik,
    Niklas Kühl
}
\affiliations {
     Karlsruhe Institute of Technology\\
    \{patrick.hemmer, michael.voessing, johannes.jakubik, niklas.kuehl\}@kit.edu, 
    lukas.thede@alumni.kit.edu
}

\begin{document}

\maketitle

\begin{abstract}
Recent research suggests that combining AI models with a human expert can exceed the performance of either alone. The combination of their capabilities is often realized by \textit{learning to defer} algorithms that enable the AI to learn to decide whether to make a prediction for a particular instance or defer it to the human expert. However, to accurately learn which instances should be deferred to the human expert, a large number of expert predictions that accurately reflect the expert's capabilities are required---in addition to the ground truth labels needed to train the AI. This requirement shared by many learning to defer algorithms hinders their adoption in scenarios where the responsible expert regularly changes or where acquiring a sufficient number of expert predictions is costly. In this paper, we propose a three-step approach to reduce the number of expert predictions required to train learning to defer algorithms. It encompasses (1) the training of an embedding model with ground truth labels to generate feature representations that serve as a basis for (2) the training of an expertise predictor model to approximate the expert's capabilities. (3) The expertise predictor generates artificial expert predictions for instances not yet labeled by the expert, which are required by the learning to defer algorithms. We evaluate our approach on two public datasets. One with ``synthetically'' generated human experts and another from the medical domain containing real-world radiologists' predictions. Our experiments show that the approach allows the training of various learning to defer algorithms with a minimal number of human expert predictions. Furthermore, we demonstrate that even a small number of expert predictions per class is sufficient for these algorithms to exceed the performance the AI and the human expert can achieve individually.
\end{abstract}

\section{Introduction}
\label{subsec:introduction}

Recent advances in the field of artificial intelligence (AI) have improved the accuracy of AI models to a point where they exceed the performance of human experts on an increasing number of tasks \cite{farrell2021identifying, Rajpurkar_CheXNetRP_2017, Topol_HighperformanceMT_2019}. However, various application domains remain where AI models can not consistently outperform human experts \cite{cremer_deeplimitations_2021, raghu_algorithmic_2019,kuhl2022human}. Motivated by this observation, prior work has started to explore how the capabilities of human experts and AI models can be combined. In this context, \textit{learning to defer} algorithms have demonstrated promising results by enabling the AI model not to make a prediction for all instances but rather to learn to defer a subset of them to the human expert---based on the AI model's and the expert's capabilities \cite{monzannar_consistent_2020,okati_differentiable_2021,raghu_algorithmic_2019}. Non-identical capabilities of the human expert and the AI can originate, e.g., from limited training data, limited model capacity, or the availability of side information only accessible to the human expert \cite{hemmer2022effect,charusaie2022sample}. By compensating for the weaknesses of both the human expert and the AI through effective instance allocation, learning to defer algorithms can achieve a higher performance level than either the human or the AI model can achieve individually \cite{raghu_algorithmic_2019,wilder_learningtocomplement_2020}.

However, large amounts of labeled data are required to train learning to defer algorithms. In fact, they require not only ground truth labels to train the AI model but also additional human expert predictions that accurately reflect the expert's capabilities \cite{leitao2022human}. In this context, any new human expert aiming to collaborate with the AI model would have to provide expert predictions so that the learning to defer algorithms can understand the expert's individual capabilities. It can become particularly costly in application domains with frequently changing human experts or where data labeling is time-consuming and requires the knowledge of highly trained specialists, e.g., in medicine. Due to this impediment, the usage of learning to defer algorithms can become infeasible in these cases---despite their potential for decision-making tasks with high costs of errors. 

To reduce the number of human expert predictions required for the training of learning to defer algorithms, we propose a novel three-step approach capable of generating artificial expert predictions from only a small number of human expert predictions accurately reflecting the expert's capabilities. (1) We train an embedding model with ground truth labels to extract feature representations which (2) serve as a basis for training an expertise predictor model that learns to approximate the human expert's capabilities with only a small number of human expert predictions. It is trained on the available expert predictions while simultaneously leveraging instances for which no human expert predictions are available by drawing upon an interchangeable semi-supervised learning component. (3) The expertise predictor generates artificial expert predictions for instances not labeled by the human expert. Both human and artificial expert predictions combined can then be used to train a variety of learning to defer algorithms. 

We empirically demonstrate the efficiency of our approach on the CIFAR-100 dataset \cite{Krizhevsky_LearningML_2009} using ``synthetically'' generated human expert predictions and on the NIH chest X-ray dataset \cite{NIHDataset,Wang_ChestXRay8HC_2017} that provides real-world individual radiologists' predictions. In addition, it also contains adjudicated ground truth labels provided by a panel of radiologists that serve as a ``gold standard''. For example, with our proposed approach, six expert predictions per class suffice to enable all evaluated learning to defer algorithms to outperform both the AI model and the radiologist compared to either conducting the task alone. On average, this corresponds to 98.96\% of the achieved accuracy compared to the training with a complete set of human expert predictions.

To summarize, the contribution of our work is threefold: First, we propose a novel approach that learns the capabilities of individual human experts from a small set of human expert predictions. Second, we demonstrate our approach's ability to enable state-of-the-art learning to defer algorithms to function with only a minimum of required human expert predictions while maintaining their desirable properties of achieving superior team performance. Third, we show its real-world applicability in the context of the medical domain with a dataset providing individual radiologists' predictions together with high-quality ground truth labels.

\section{Related Work}
\label{subsec:relatedWork}

Over the last years, research has investigated how human and AI capabilities can be combined to achieve superior team performance compared to both performing tasks individually \cite{madras_predict_2018,monzannar_consistent_2020,okati_differentiable_2021,raghu_algorithmic_2019,wilder_learningtocomplement_2020}. In this context, \textit{learning to defer} algorithms that enable the AI model to decide whether to make a prediction for a particular instance or defer it to a human expert have demonstrated promising results \cite{leitao2022human}. \citet{raghu_algorithmic_2019} propose to estimate the prediction confidence of the classifier and the human expert on an instance basis. Instances are deferred when the human expert is estimated to be more confident than the classifier. While this approach optimizes the classifier in isolation, \citet{wilder_learningtocomplement_2020} propose the idea of optimizing for team performance by training the classifier to specifically complement the human expert's capabilities. In this context, several approaches jointly train the classifier together with a deferral system \cite{madras_predict_2018,okati_differentiable_2021,wilder_learningtocomplement_2020}. Further work utilizes objective functions with theoretical guarantees for regression \cite{de_regression_2020} and classification tasks \cite{de_classification_2021}. Moreover, \citet{monzannar_consistent_2020} propose a consistent surrogate loss function inspired by cost-sensitive learning. \citet{raman_improving_2021} build upon their work by individualizing the approach to a particular human expert via fine-tuning. Further research studies the assignment of instances in the context of bandit feedback \cite{gao_human_2021} or suggests learning to allocate instances within teams consisting of an AI model and multiple human experts \cite{hemmerijcai,keswani_towardsunbiased_2021}.

These approaches have in common that they require human expert predictions to recognize an expert's strengths and weaknesses when deciding whether an instance shall be deferred. Since the capabilities of human experts can vary in different areas of the feature space, e.g., due to different levels of knowledge or biases \cite{Lampert2016AnES}, this requires the availability of individual predictions for each expert working with the AI model, which impedes the applicability of learning to defer algorithms, especially in application domains with frequently changing human experts or high acquisition costs for expert predictions. 

Our approach addresses this limitation through the generation of artificial expert predictions only from a small number of predictions provided by an individual human expert. Concurrent work designs an active learning scheme to reduce the number of required predictions \cite{charusaie2022sample}. On the contrary, our approach does not require iteratively identifying instances for which human expert predictions are queried. Instead, given a small number of human expert predictions, it learns to infer artificial ones for unlabeled instances in the training dataset that can then be used for the training of learning to defer algorithms.

Besides their benefits with regard to the overall team performance, research has also investigated the effect of learning to defer algorithms in the context of fairness, as AI models are known to be capable of generating biased predictions against protected groups \cite{pmlr-v81-buolamwini18a}. Particularly, when AI models can abstain from uncertain predictions, e.g., in selective classification, this can result in amplifying existing biases \cite{jones2021selective}. A similar phenomenon can also be observed in the behavior of humans since their decisions are neither free from prejudices \cite{Tversky1974}. To prevent possible amplifications of AI model and human biases, \citet{madras_predict_2018} propose a learning to defer algorithm that considers both overall team performance and fairness by introducing a regularized loss function combining the error rate with a fairness metric. They find that their approach can result in a system that is more accurate and less biased. Moreover, \citet{keswani_towardsunbiased_2021} propose to ensure that their learning to defer algorithm produces unbiased predictions by balancing error rates for all protected groups and drawing upon the minimax Pareto fairness concept.

Lastly, another related stream of work focuses on efficiently combining predictions of multiple human annotators to derive high-quality labels, e.g., from crowdsourced predictions \cite{branson2017lean,guan_who_2018,liao2021towards,welinder_online_2010}. Whereas these approaches focus on ground truth label quality using a larger worker pool, we are interested in learning the capabilities of an individual human expert only from a minimal number of expert predictions.

\section{Problem Formulation}
\label{subsec:problemstatement}

In this section, we consider the learning to defer setting for classification tasks. Given an instance \(x_i \in \mathcal{X}\), the goal is to learn to predict its ground truth label \( y_i \in \mathcal{Y} = \{1, \ldots, k\} \) with \(k\) denoting the number of classes. The indices of the respective instances are denoted as $i \in N = \{1, ..., n\}$. In addition to \(y_i\), we assume to have access to the prediction of a human expert \(h_i \in \mathcal{Y}\). The expert's predictions can deviate from the ground truth as individual human experts' capabilities vary in different areas of the feature space. Possible reasons can be different levels of background knowledge, experience, or biases \cite{Lampert2016AnES}. Moreover, ground truth labels are often curated in various application domains through multiple human experts to ensure high label quality \cite{guan_who_2018}. For example, in the medical domain, a panel of radiologists is often involved in determining the prevalence of a certain clinical finding \cite{NIHDataset, Wang_ChestXRay8HC_2017}. In this context, these experts most likely differ from the ones that actually cooperate with the deployed AI model in practice.

As it is often accompanied by high costs to acquire a large number of individual human expert predictions, particularly when the expert that collaborates with the AI model changes frequently, we assume individual human expert predictions to be available only for a small subset of instances \(L \subseteq N\) with \(|L| \ll |N|\). For the remaining subset \(U = N \setminus L\), no individual human expert predictions are available. Hence, we define the dataset with ground truth labels and human expert predictions as \(D^l = {\{(x_i, y_i, h_i)\}}^l_{i=1}\) with \(l = |L|\). Similarly, we define the dataset without such predictions but with ground truth labels as \(D^u = {\{(x_j, y_j)\}}^u_{j=1}\) with \(u = |U|\) such that $D^l \cap D^u = \emptyset$. 

The general idea of learning to defer algorithms is to train a deferral system \(A: \mathcal{X} \rightarrow \{0,1\}\) that decides whether to defer an instance to the human expert, with \(A(x_i) = 1\) denoting the deferral decision and \(A(x_i) = 0\) referring to make a prediction using a classifier \(F: \mathcal{X} \rightarrow \mathcal{Y}\). However, jointly optimizing the deferral system \(A\) and the classifier \(F\) requires the availability of a human expert's prediction \(h_i\) and ground truth label \(y_i\) for each instance \(x_i\) which is only met for the small subset \(D^l\). With our proposed approach, we learn to generate an artificial expert prediction \(\hat{h}_j \in \mathcal{Y}\) for each instance in \(D^u\) that accurately reflects the expert's capabilities and enables the training of \(A\) and \(F\) using all instances from $D = D^l \cup D^u$. Keeping the approach independent in the form of a prior step makes it applicable to any learning to defer algorithm requiring both human expert predictions and ground truth labels for the training.

\section{Approach}
\label{subsec:approach}
In this section, we describe our proposed approach for learning the capabilities of an individual human expert to generate artificial expert predictions. It consists of three steps: (1) Training of an \textit{embedding model} with ground truth labels, which is used to extract feature representations. (2) They serve as input for the training of an \textit{expertise predictor model} to approximate the human expert's capabilities. (3) The expertise predictor model generates \textit{artificial expert predictions} for the instances not labeled by the human expert. Both human and artificial expert predictions can then be used for the training of learning to defer algorithms.

\subsection{Training of Embedding Model}

The first component of the approach embeds each instance into a feature representation using a feature extractor. We denote the model as $\Phi_{emb}: \mathcal{X} \rightarrow \mathbb{R}^d$ with $d$ being the length of the feature vectors. The embedding model is trained together with a classifier $\Omega: \mathbb{R}^d \rightarrow \mathcal{Y}$. The classifier uses the feature representations of the embedding model to learn the classification task based on all data instances of $D$ and their respective ground truth labels. The embedding model is trained using the loss:
\begin{equation}
    \mathcal{L} = \frac{1}{n} \sum_{i=1}^n H(y_i, \Omega(\Phi_{emb}(x_i)))
\end{equation}
with \(H\) being the cross-entropy between a ground truth label and the classifier's prediction. Using the trained embedding model \(\Phi_{emb}\), we can transpose each instance \(x_i\) into a feature vector \(f_i \in \mathbb{R}^d\ \forall i \in N\) thereby constricting the hypothesis space for the subsequent task of approximating the human expert's capabilities. This is an approach commonly used in few-shot learning \cite{koch_siameseNN_2015,scott_adapted_2018, snell_prototypical_2017,tian_rethinking_2020,vinyals_matching_2017} to transfer prior knowledge from the task of predicting ground truth labels---to the novel task of generating artificial expert predictions.

\subsection{Training of Expertise Predictor Model} 
The second component is the expertise predictor model. It learns to approximate the human expert's capabilities reflected in the expert's labeling behavior. For a multi-class classification task with \(k > 2\), there are \(k-1\) possibilities for an incorrect prediction. In case a human expert does not systematically confound certain classes when making incorrect predictions, identifying the exact class of an incorrect prediction can make the task of learning the expert's labeling behavior unnecessarily complex. Therefore, we propose to reduce the complexity to a binary task---learning whether the expert makes a correct or incorrect prediction. We use the following function to translate a multi-class expert prediction \(h_i\) into a binary expert prediction \(h^{bin}_i\):
\begin{equation} \label{eq:bin_expert_label}
    \Psi(h_i, y_i) = \begin{cases} 1, & \text{if } h_i = y_i \\ 0, & \text{otherwise} \end{cases},  \quad \forall i \in L.
\end{equation}
The resulting binary expert predictions \(h^{bin}_i\) are then used together with the feature representations \(f_i\ \forall i \in L\) as input to train the expertise predictor model \(\Phi_{ex}: \mathbb{R}^d \rightarrow \{0,1\}\).
\begin{figure*}[t]
    \centering
    \includegraphics[width=0.98\linewidth]{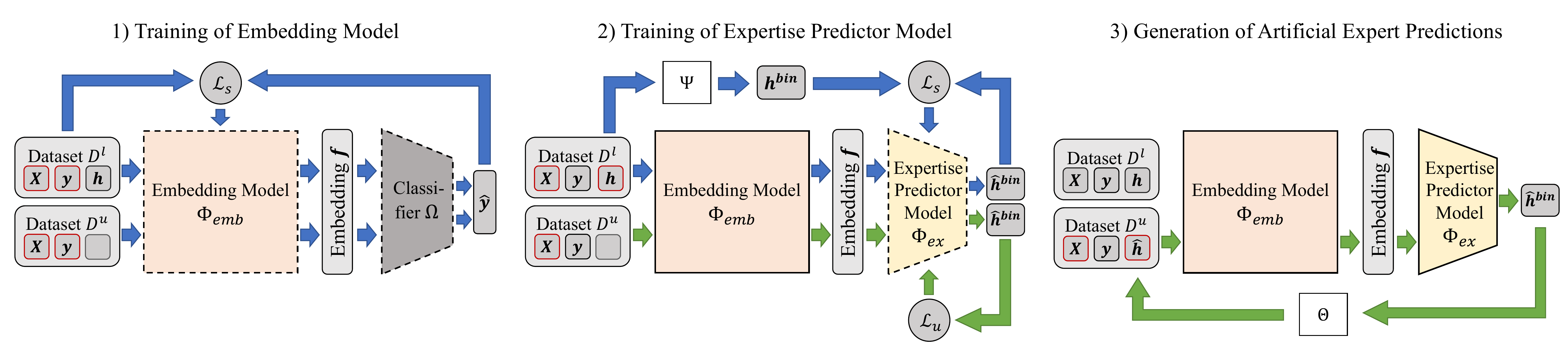}
    \caption{The proposed approach for learning the capabilities of a human expert. We use blue arrows to visualize the path of supervised instances, while green arrows represent unsupervised instances. Red borders denote the relevant components of the dataset \(D^l\) and \(D^u\) that are used at the respective step.
    }
    \label{fig:framework_training_graph}
\end{figure*}

In order to leverage both labeled and unlabeled instances for this task, we propose to incorporate an interchangeable semi-supervised learning component for training the expertise predictor model. We instantiate this component with two state-of-the-art semi-supervised learning algorithms to demonstrate its flexibility. In detail, we use FixMatch and CoMatch as both have demonstrated that they can outperform other semi-supervised approaches \cite{sohn_fixmatch_2020,li_comatch_2021}. We denote these two instantiations of the proposed approach as \textit{Embedding-FixMatch} and \textit{Embedding-CoMatch} respectively. Both algorithms combine methods from consistency regularization and entropy minimization. The loss function of both algorithms is displayed in \Cref{eq:loss_fixmatch-comatch}. Whereas FixMatch's loss function consists of a supervised and an unsupervised loss term, CoMatch includes an additional contrastive loss term:
\begin{equation}\label{eq:loss_fixmatch-comatch}
\begin{split}
\mathcal{L} &= \underbrace{\frac{1}{l} \sum_{i=1}^l H(h^{bin}_i, \Phi_{ex}(\Phi_{emb}(\text{Aug}_w(x_i))))}_{\mathcal{L}_s} + \\ 
\lambda_{u} &* \underbrace{\frac{1}{u} \sum_{j=1}^u \mathbb{1}{(\max(q_j) \geq \tau)} H(\hat{q}_j, \Phi_{ex}(\Phi_{emb}(\text{Aug}_s(x_j))))}_{\mathcal{L}_u} + \\
\lambda_{c} &* \underbrace{\frac{1}{u} \sum_{j=1}^u H(\hat{W}^q_j,\hat{W}^z_j)}_{\mathcal{L}_c \text{ (CoMatch only)}}.
\end{split}
\end{equation}
In both cases, the first part of the loss \(\mathcal{L}_s\) represents the cross-entropy \(H\) between a binary expert prediction \(h^{bin}_i\) and the model's prediction for a weakly augmented input \(\text{Aug}_w(x_i)\) with \(i \in L\). For the second part \(\mathcal{L}_{u}\), the predicted class distribution of the model \( q_j = \Phi_{ex}(\Phi_{emb}(\text{Aug}_w(x_j)))\) is computed given an unlabeled weakly augmented image \(x_j\) with \(j \in U\). FixMatch calculates \(\hat{q}_j = \arg \max (q_j)\) to obtain a hard pseudo-label. The cross-entropy compares the pseudo-label against the model prediction for a strong augmentation of the same instance \(\text{Aug}_s(x_j)\), with \(\tau\) referring to a threshold above which the pseudo-label is retained. In contrast, CoMatch does not convert \(q_j\) to a hard pseudo-label \(\hat{q}_j\) but performs memory-smoothed pseudo-labeling to obtain a pseudo-label. For the third part \(\mathcal{L}_{c}\), CoMatch constructs a normalized pseudo-label graph \(\hat{W}^q\) and a normalized embedding graph \(\hat{W}^z\). The first graph measures the similarity of instances in the label space. It is used as the target for the training of an embedding graph that quantifies the similarity of strongly-augmented instances in the embedding space. For the embedding graph, CoMatch requires the expertise predictor model to have an additional projection head transforming the feature representation \(f_j\) into a low-dimensional embedding via two fully connected layers \cite{li_comatch_2021}. Both graphs are used to calculate a contrastive loss minimizing the cross-entropy between them for all unlabeled instances \(j \in U\). Lastly, the second and third components of the loss are controlled by the fixed hyperparameters \(\lambda_{u}\) and \(\lambda_{c}\) respectively. For further technical details, we refer to \citet{sohn_fixmatch_2020} for FixMatch and \citet{li_comatch_2021} for CoMatch.

\subsection{Generation of Artificial Expert Predictions} 
After the training of both components, we obtain an artificial binary expert prediction \(\hat{h}^{bin}_j\) by transposing \(x_j\) into a feature representation \(f_j\) using the embedding model together with the expertise predictor model \(\hat{h}^{bin}_j = \arg \max \Phi_{ex}(\Phi_{emb}(x_j))\ \forall j \in U\). Finally, we translate all artificial binary expert predictions \(\hat{h}^{bin}_j\) into multi-class artificial expert predictions \(\hat{h}_j\ \forall j \in U\) in the following way:
\begin{equation} \label{eq:mult_expert_labels}
    \Theta(\hat{h}^{bin}_j, y_j) = \begin{cases} y_j, & \text{if } \hat{h}^{bin}_j = 1 \\ p, & \text{otherwise}\end{cases}, p \sim \mathcal{U}(\{1,\ldots, k\} \setminus y_j).
\end{equation}
\Cref{fig:framework_training_graph} provides an overview of the proposed approach with the FixMatch loss terms as an exemplary instantiation of the semi-supervised learning component. We formalize the approach in Algorithm A1 in the Appendix.
\begin{table*}[h]
  \centering
  \resizebox{\textwidth}{!}{
  \begin{tabular}{lllllllll}
    \toprule
    &\multicolumn{1}{c}{$l$} & \multicolumn{1}{c}{40} & \multicolumn{1}{c}{80} & \multicolumn{1}{c}{120} & \multicolumn{1}{c}{200} & \multicolumn{1}{c}{400} & \multicolumn{1}{c}{1000} & \multicolumn{1}{c}{5000} \\
    \midrule
    \multirow{6}*{$H_{60}$} & FixMatch & 68.81 ($\pm$0.51) & 71.72 ($\pm$1.73) & 74.93 ($\pm$2.42) & 75.42 ($\pm$2.00) & 78.57 ($\pm$1.90) & 79.62 ($\pm$1.72) & 82.67 ($\pm$0.44)\\
    & CoMatch & 75.00 ($\pm$2.49) & 75.97 ($\pm$1.20) & 79.17 ($\pm$2.32) & 81.46 ($\pm$1.36) & 84.61 ($\pm$0.86) & 85.33 ($\pm$0.67) & 87.62 ($\pm$0.51)\\
    & Embedding-NN & 75.74 ($\pm$1.57) & 77.57 ($\pm$1.73) & 79.32 ($\pm$1.65) & 79.46 ($\pm$2.08) & 80.78 ($\pm$0.49) & 81.47 ($\pm$0.24) & 83.77 ($\pm$0.13)\\
    & Embedding-SVM & 74.49 ($\pm$1.48) & 79.49 ($\pm$1.49) & 81.61 ($\pm$1.31) & 81.81 ($\pm$1.61) & 84.69 ($\pm$1.15) & \textbf{86.41} ($\pm$0.37) & \textbf{89.16} ($\pm$0.33)\\
    \cmidrule(r){2-9} 
    & \textbf{Embedding-FixMatch} & \textbf{79.69} ($\pm$0.02) & 80.52 ($\pm$0.04) & \textbf{84.06} ($\pm$0.36) & \textbf{83.82} ($\pm$1.15) & \textbf{85.50} ($\pm$0.07) & 86.04 ($\pm$0.34) & 86.95 ($\pm$0.15)\\
    & \textbf{Embedding-CoMatch} & 79.65 ($\pm$0.02) & \textbf{80.88} ($\pm$0.02) & 84.01 ($\pm$0.27) & 83.74 ($\pm$1.17) & 85.38 ($\pm$0.14) & 85.95 ($\pm$0.36) & 86.87 ($\pm$0.05)\\
    \midrule
    \multirow{6}*{$H_{90}$} & FixMatch & 89.51 ($\pm$1.16) & 90.23 ($\pm$0.72) & 91.44 ($\pm$0.92) & 93.21 ($\pm$0.63) & 93.88 ($\pm$0.26) & 94.65 ($\pm$0.46) & 95.84 ($\pm$0.06)\\
    & CoMatch & 87.24 ($\pm$2.02) & 90.09 ($\pm$0.75) & 91.72 ($\pm$0.85) & 92.82 ($\pm$0.45) & 93.69 ($\pm$0.64) & 95.15 ($\pm$0.32) & \textbf{96.62} ($\pm$0.12)\\
    & Embedding-NN & 92.10 ($\pm$0.82) & 92.35 ($\pm$0.67) & 92.70 ($\pm$0.65) & 92.86 ($\pm$0.79) & 93.48 ($\pm$0.22) & 93.63 ($\pm$0.22) & 93.68 ($\pm$0.20)\\
    & Embedding-SVM & 92.08 ($\pm$0.14) & 93.00 ($\pm$0.62) & 93.46 ($\pm$0.51) & 93.78 ($\pm$0.52) & 94.21 ($\pm$0.35) & 95.53 ($\pm$0.29) & 96.38 ($\pm$0.24)\\
    \cmidrule(r){2-9} 
    & \textbf{Embedding-FixMatch} & 92.44 ($\pm$0.01) & 93.82 ($\pm$0.17) & 95.34 ($\pm$0.02) & 95.02 ($\pm$0.02) & \textbf{95.70} ($\pm$0.02) & \textbf{95.81} ($\pm$0.02) & 96.25 ($\pm$0.02)\\
    & \textbf{Embedding-CoMatch} & \textbf{94.75} ($\pm$0.02) & \textbf{95.10} ($\pm$0.01) & \textbf{95.46} ($\pm$0.01) & \textbf{95.11} ($\pm$0.01) & 95.58 ($\pm$0.01) & 95.63 ($\pm$0.01) & 96.14 ($\pm$0.02)\\
    \bottomrule
  \end{tabular}}
    \caption{\(F_{0.5}\)-score (mean and standard deviation) of the artificial binary expert predictions generated by the expertise predictor model using different numbers of \(l\) available human expert predictions for the synthetic experts \(H_{60}\) and \(H_{90}\) on CIFAR-100.}\label{table:results_predex_cifar}
\end{table*}

\section{Experiments}
\label{subsec:experiments}
We present the experimental evaluation of our proposed approach on CIFAR-100 \cite{Krizhevsky_LearningML_2009} with synthetic human expert predictions and on the NIH dataset \cite{NIHDataset,Wang_ChestXRay8HC_2017} with real-world radiologists' predictions. Further implementation details and results are presented in the Appendix, which we provide together with the code at \url{https://github.com/ptrckhmmr/learning-to-defer-with-limited-expert-predictions}.

\subsection{Experimental Setup}
The performance of the approach is evaluated in two steps. First, we evaluate how well the artificial binary expert predictions generated by the expertise predictor model represent the true binary predictions of the human experts. Second, we assess the performance difference of multiple learning to defer algorithms trained with both human and artificial expert predictions compared to the training with the entire set of human expert predictions. Both steps are evaluated with different numbers of instances for which \(l = |D^l|\) human expert predictions are provided to learn the expert's capabilities with \(l = m*k\) and \(m \in \{2, 4, 6, 10, 20, 50, 250\}\).

\paragraph{Evaluation of Artificial Expert Predictions.} 
For each \(l\), we draw instances randomly from the training data while ensuring balanced class proportions. We use the $F_{0.5}$-score as the metric to evaluate the performance of the artificial binary expert predictions, as the transformation according to \Cref{eq:bin_expert_label} can result in an imbalanced class distribution with regard to whether the human expert provides a correct or incorrect prediction.

\paragraph{Evaluation of Learning to Defer Algorithms.} 
As our approach can be applied independently as a prior step to learning to defer algorithms, we demonstrate its applicability to the approaches of \citet{monzannar_consistent_2020}, \citet{raghu_algorithmic_2019}, and \citet{okati_differentiable_2021}. \citet{monzannar_consistent_2020} jointly train the classification and deferral decision with a consistent surrogate loss function by adding an additional deferral class to the classification problem. \citet{raghu_algorithmic_2019} train a classifier and a human error model separately. Each instance is allocated by comparing both models' confidences. \citet{okati_differentiable_2021} propose to assign each instance via a negative log-likelihood loss calculated on an instance basis for both the classifier and expert. The classifier is then trained to complement the human expert by minimizing the negative log-likelihood loss only on the instances assigned to the classifier.  

As both the CIFAR-100 and the NIH datasets are balanced across classes, we use the system accuracy to evaluate the performance of the learning to defer algorithms trained on both the available human expert predictions \(l\) and the artificial expert predictions. For each \(l\), we report their performances relative to the performance of the respective approach trained on a complete set of expert predictions \(|D^l|=|D| \text{ and } D^u = \emptyset\) (\textit{Complete Expert Predictions}), which serves as upper boundary. Furthermore, we consider the performance of the human expert and the AI alone as two lower boundaries (\textit{Human Expert Alone} and \textit{Classifier Alone}) that have to be exceeded to demonstrate our approach's applicability to learning to defer algorithms. 

\paragraph{Baselines.} 
We benchmark our proposed embedding semi-supervised learning (\textit{Embedding-SSL}) approaches \textit{Embedding-FixMatch} and \textit{Embedding-CoMatch} against the following baselines: First, we use the algorithms \textit{FixMatch} \cite{sohn_fixmatch_2020} and \textit{CoMatch} \cite{li_comatch_2021} without an embedding model as semi-supervised learning baselines (\textit{SSL}). Second, we consider two embedding supervised learning (\textit{Embedding-SL}) baselines that use an embedding model combined with a supervised classifier as an expertise predictor model. Specifically, we implement an \textit{Embedding-NN} baseline, which uses a single-layer neural network as an expertise predictor model, and an \textit{Embedding-SVM} baseline, where the expertise predictor model consists of a support vector classifier.

\subsection{CIFAR-100}\label{subsec:cifar_100_dataset}

\begin{figure*}[t]
    \centering
    \includegraphics[width=0.9\linewidth]{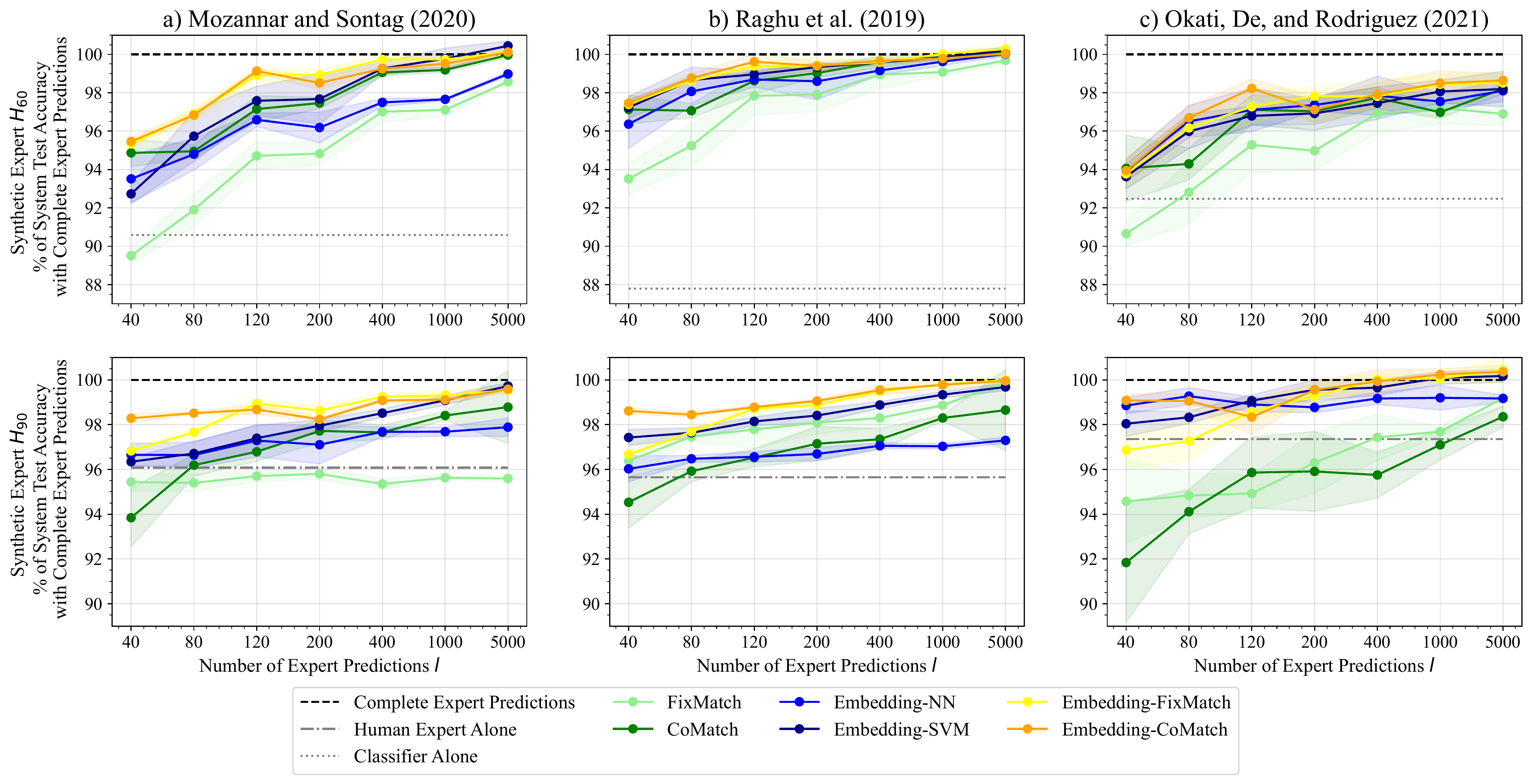}
    \caption{Experimental results for the learning to defer algorithms of a) \citet{monzannar_consistent_2020}, b) \citet{raghu_algorithmic_2019}, and c) \citet{okati_differentiable_2021} trained on different numbers of human expert predictions \(l\) and artificial expert predictions for \(H_{60}\) and \(H_{90}\) on CIFAR-100. Artificial expert predictions are generated using different numbers of human expert predictions \(l\). We report mean and standard deviation over five seeds and omit the lower boundary for better visualization.}
    \label{fig:human_machine_results_cifar}
\end{figure*}

\paragraph{Dataset and Expert Generation.} The CIFAR-100 dataset \cite{Krizhevsky_LearningML_2009} consists of 60,000 images from 100 subclasses which are grouped into 20 superclasses of equal size. We use the 20 superclasses for the classification task and the 100 subclasses for the generation of synthetic human experts. This allows us to model the expert's capabilities on a fine-granular level. To model a synthetic expert, by design, we draw a subset of the 100 subclasses as the expert's strengths, i.e., subclasses for which the expert's predictions are correct. The remaining subclasses are defined as the expert's weaknesses, i.e., subclasses for which the expert's predictions are incorrect. The subset of the expert's strengths subclasses is defined by selecting one of the subclasses uniformly at random as a strength base. The remaining strength subclasses are drawn randomly, weighted by their similarity to the strength base. We calculate the similarities between the subclasses as the average cosine distance between the feature representations of the instances, which are obtained from an EfficientNet-B1 model \cite{tan_efficientnet_2020} pretrained on ImageNet. For the remaining weakness subclasses, the expert predicts randomly across all subclasses weighted by the similarity to the true subclass. Finally, the respective subclasses are mapped to the corresponding superclasses for the actual classification task.

\paragraph{Implementation.}\label{subsubsec:implementation} 
We allocate 40,000 images to the training and 10,000 images to the validation split while reserving 10,000 images for the test split. As an embedding model, we use an EfficientNet-B1 \cite{tan_efficientnet_2020}. The expertise predictor model consists of a single fully connected layer with a softmax activation. For the learning to defer algorithms, we use a WideResNet \cite{Zagoruyko_WideRN_2016} as backbone. We consider a strong synthetic expert \(H_{90}\) (90 of 100 subclasses as strengths) as well as a weaker one \(H_{60}\) (60 of 100 subclasses as strengths) and evaluate all approaches with the following number of available expert predictions: \(l \in \{40, 80, 120, 200, 400, 1000, 5000\}\). We train the embedding model for \(200\) epochs using SGD as an optimizer with Nesterov momentum and a learning rate of 0.1. Each expertise predictor model of our \textit{Embedding-SSL} approaches is trained for \(50\) epochs using SGD with a learning rate of 0.03. We repeat each experiment five times with different random seeds.

\paragraph{Artificial Expert Prediction Results.} 
We report the performances of the artificial binary expert predictions of the expertise predictor model against the binary expert predictions for both synthetic experts in \Cref{table:results_predex_cifar}. Overall, the proposed \textit{Embedding-SSL} approaches outperform all baselines for \(l < 1,000\). More specifically, the results show that for \(l < 1,000\) our proposed \textit{Embedding-SSL} approaches outperform the \textit{SSL} baselines, on average, by 8.13\% (for \(H_{60}\)) and 3.80\% (for \(H_{90}\)). Moreover, the proposed \textit{Embedding-SSL} approaches are able to outperform the \textit{Embedding-SL} baselines for \(l < 1,000\), on average, by 4.09\% (for \(H_{60}\)) and 1.97\% (for \(H_{90}\)). This demonstrates the benefit of using semi-supervised learning within the embedding space.
\begin{table*}[h]
  \centering
  \resizebox{\linewidth}{!}{\begin{tabular}{llllllll}
    \toprule
    \multicolumn{1}{c}{$l$}     & \multicolumn{1}{c}{4} & \multicolumn{1}{c}{8} & \multicolumn{1}{c}{12} & \multicolumn{1}{c}{20} & \multicolumn{1}{c}{40} & \multicolumn{1}{c}{100} & \multicolumn{1}{c}{500} \\
    \midrule
    FixMatch & 56.97 ($\pm$7.26) & 59.43 ($\pm$4.68) & 72.63 ($\pm$4.43) & 62.09 ($\pm$3.62) & 56.39 ($\pm$3.43) & 83.40 ($\pm$1.34) & 87.25 ($\pm$1.05)\\
    CoMatch & 70.88 ($\pm$5.50) & 69.17 ($\pm$3.45) & 76.76 ($\pm$2.01) & 67.86 ($\pm$1.07) & 69.79 ($\pm$1.92) & 84.45 ($\pm$1.42) & 89.77 ($\pm$0.33)\\
    Embedding-NN & 85.87 ($\pm$4.72) & 88.61 ($\pm$1.75) & 89.03 ($\pm$1.44) & 91.14 ($\pm$0.91) & 91.35 ($\pm$1.37) & 91.27 ($\pm$0.52) & 92.29 ($\pm$0.53)\\
    Embedding-SVM & 59.86 ($\pm$33.44) & 86.34 ($\pm$1.63) & 86.89 ($\pm$2.32) & 90.27 ($\pm$0.95) & 91.49 ($\pm$0.98) & 92.07 ($\pm$0.55) & \textbf{92.74} ($\pm$0.96)\\
     \midrule
    \textbf{Embedding-FixMatch } & \textbf{90.74} ($\pm$0.12) & \textbf{92.29} ($\pm$0.00) & 92.56 ($\pm$0.00) & \textbf{91.65} ($\pm$0.03) & \textbf{92.69} ($\pm$0.13) & \textbf{92.22} ($\pm$0.08) & 91.96 ($\pm$0.17)\\
    \textbf{Embedding-CoMatch } & 88.56 ($\pm$1.34) & 92.28 ($\pm$0.12) & \textbf{92.63} ($\pm$0.11) & 91.57 ($\pm$0.20) & 90.84 ($\pm$0.44) & 91.93 ($\pm$0.00) & 91.07 ($\pm$0.20)\\
     \bottomrule
  \end{tabular}}
    \caption{\(F_{0.5}\)-score (mean and standard deviation) of the artificial expert predictions generated by the expertise predictor model using different numbers of \(l\) available human expert predictions of radiologist 4295342357 on the NIH dataset.}\label{table:results_predex_nih}
\end{table*}
\begin{figure*}[h]
    \centering
    \includegraphics[width=0.9\linewidth]{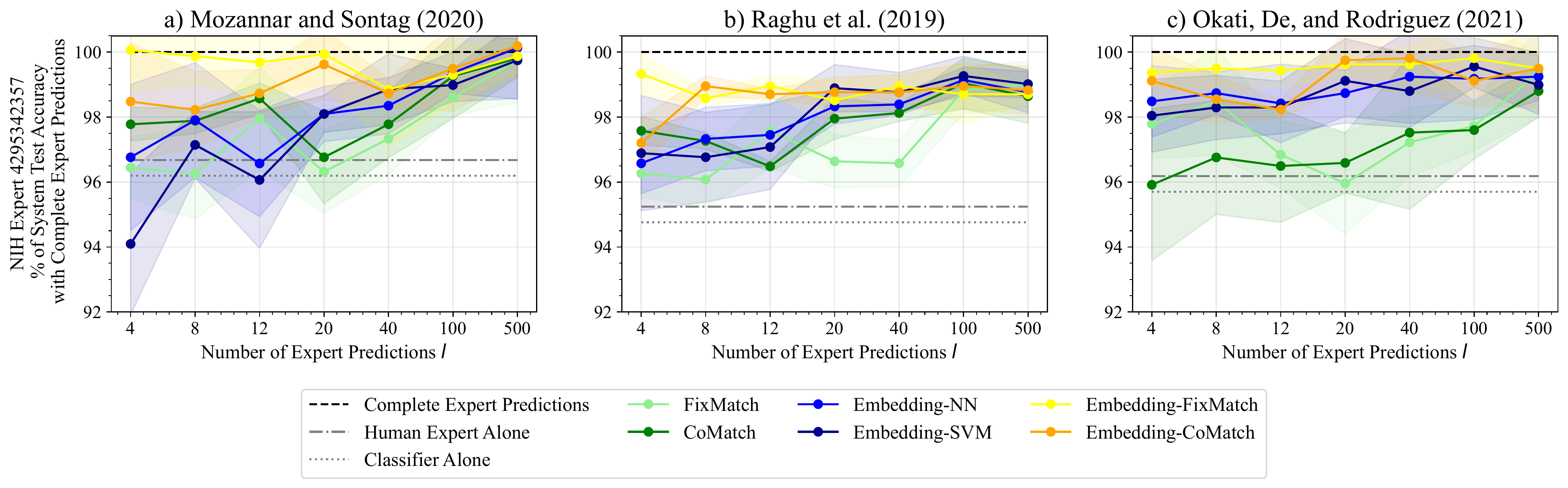}
    \caption{Experimental results for the learning to defer algorithms of a) \citet{monzannar_consistent_2020}, b) \citet{raghu_algorithmic_2019}, and c) \citet{okati_differentiable_2021} trained on different numbers of human expert predictions \(l\) and artificial expert predictions for radiologist 4295342357. Artificial expert predictions are generated using different numbers of human expert predictions \(l\). We report mean and standard deviation over five seeds.}
    \label{fig:human_machine_results_nih}
\end{figure*}

\paragraph{Learning to Defer Results.} 
\Cref{fig:human_machine_results_cifar} visualizes the results of the experiments with the approaches of \citet{monzannar_consistent_2020}, \citet{raghu_algorithmic_2019}, and \citet{okati_differentiable_2021}. They show the general feasibility of training learning to defer algorithms using artificial expert predictions generated by our proposed approach, as all three outperform the individual team members in all experiments. Furthermore, the results indicate that even with only a few human expert predictions, the performances of the learning to defer algorithms converge quickly towards the upper boundary, matching the accuracy achieved with a fully labeled dataset. Regarding the approach of \citet{monzannar_consistent_2020}, the \textit{Embedding-SSL} approaches reach, on average, 99.02\% (for \(H_{60}\)) and 98.81\% (for \(H_{90}\)) of the upper boundary with only six expert predictions per class. Furthermore, the results support the previous statements concerning the superiority of the \textit{Embedding-SSL} approaches. For \(l<400\), they outperform the \textit{SSL} baselines, on average, by 3.27\% (for \(H_{60}\)) and 2.47\% (for \(H_{90}\)), and the \textit{Embedding-SL} baselines, on average, by 1.99\% (for \(H_{60}\)) and 1.25\% (for \(H_{90}\)). The results of the approach of \citet{raghu_algorithmic_2019} confirm the advantage of the \textit{Embedding-SSL} approaches. For \(l<400\), they improve the \textit{SSL} baselines, on average, by 1.78\% (for \(H_{60}\)) and 1.68\% (for \(H_{90}\)). Regarding the \textit{Embedding-SL}, an average improvement of 0.53\% (for \(H_{60}\)) and 1.22\% (for \(H_{90}\)) can be achieved for \(l<400\). Lastly, for the approach of \citet{okati_differentiable_2021}, the \textit{Embedding-SSL} approaches outperform the \textit{SSL} baselines, on average, by 1.96\% (for \(H_{60}\)) and 3.92\% (for \(H_{90}\)) for \(l < 400\). Regarding the \textit{Embedding-SL} baselines, our \textit{Embedding-SSL} approaches achieve an average improvement of 0.40\% (for \(H_{60}\)) and 0.06\% (for \(H_{90}\)) over all $l$.

\subsection{NIH Chest X-rays}\label{subsec:nih_chest_x_rays}

\paragraph{Dataset.} 
The NIH dataset consists of radiologists’ predictions for four radiographic findings on 4,374 chest X-ray images from the ChestX-ray8 dataset \cite{NIHDataset,Wang_ChestXRay8HC_2017}. Contrary to the previous experiment, where we induced a synthetic connection between certain features, the degree to which the features are associated with experts' capabilities remains ex-ante unclear for the NIH dataset. Each image is labeled by a panel of three radiologists denoting the ``gold standard'' ground truth labels. Moreover, the predictions of each of the three individual radiologists are reported for each image. We select the radiologist that provides the largest number of expert predictions (labeler-id: 4295342357) and focus on the diagnosis of the clinical finding \textit{airspace opacity} for our experiments due to its balanced prevalence in the dataset. The experimental results using a further radiologist are reported in the Appendix.

\paragraph{Implementation.} 
We split the 2,350 available images into a training and test set by randomly selecting 20\% of the patients as test patients. For the embedding model and the learning to defer algorithms, we use a ResNet18 \cite{He_DeepRL_2016} pretrained on the CheXpert dataset---a different X-ray dataset \cite{Irvin_CheXpertAL_2019}. The expertise predictor model consists of a single fully connected layer with a softmax activation. We consider the following numbers of expert predictions: $l \in \{4, 8, 12,20, 40, 100, 500\}$. We train the embedding model for 200 epochs using SGD as an optimizer with Nesterov momentum and a learning rate of 0.001. The expertise predictor models of our \textit{Embedding-SSL} approaches are trained for 25 epochs using SGD with a learning rate of 0.03. We repeat the experiment five times with different random seeds.

\paragraph{Artificial Expert Prediction Results.} 
The results of the artificial expert predictions are reported in \Cref{table:results_predex_nih}. They demonstrate that, on average, the proposed \textit{Embedding-SSL} approaches outperform both \textit{SSL} baselines by 38.83\% for \(l < 100\). This confirms the result of the experiments conducted on CIFAR-100 regarding the impact of the embedding model. Furthermore, regarding the \textit{Embedding-SL} baselines, the proposed \textit{Embedding-SSL} approaches achieve an average performance improvement of 7.04\% for \(l<100\), while for \(l \geq 100\), the \textit{Embedding-SL} baselines are on par with the performances of the proposed approaches.

\paragraph{Learning to Defer Results.} 
\Cref{fig:human_machine_results_nih} displays the results of the learning to defer algorithms trained on human and artificial expert predictions for different numbers of available human expert predictions on the NIH dataset. These results underline the feasibility of using artificial expert predictions generated by the \textit{Embedding-SSL} approaches as the respective system accuracies exceed the lower baselines for all \(l\). For the algorithm of \citet{monzannar_consistent_2020}, the proposed \textit{Embedding-SSL} approaches outperform the \textit{SSL} baselines for \(l < 100\), on average, by 1.97\%. Compared to the \textit{Embedding-SL} baselines, our \textit{Embedding-SSL} approaches achieve an average improvement of 2.10\% for \(l < 100\). A closer look at the algorithm of \citet{raghu_algorithmic_2019} reveals that the proposed \textit{Embedding-SSL} approaches outperform the \textit{SSL} and \textit{Embedding-SL} baselines, on average, by 1.67\% and 1.06\% for \(l < 100\) respectively. For the algorithm of \citet{okati_differentiable_2021}, the \textit{Embedding-SSL} approaches improve the \textit{SSL} baselines, on average, by 2.42\% and the \textit{Embedding-SL} baselines by 0.69\% for \(l < 100\). For example, with only six expert predictions per class, the \textit{Embedding-SSL} approaches enable the algorithm of \citet{monzannar_consistent_2020} to achieve, on average, 99.12\% and the ones of \citet{raghu_algorithmic_2019} and \citet{okati_differentiable_2021} to achieve 98.83\% of the performance compared to when they are trained with all available expert predictions.

\paragraph{Bias.} Approximating the capabilities of a human expert through artificial expert predictions can entail the risk of inducing or amplifying prediction performance disparities between groups within a population. Therefore, for all learning to defer algorithms, we evaluate whether our proposed approach amplifies accuracy disparities with respect to patients' gender and age. For each number of human expert predictions \(l\) used to generate artificial ones, we calculate bias as the absolute difference in the accuracy between male and female patients. We can then compare it with the bias of the upper boundary (\textit{Complete Expert Predictions}). We also apply this procedure to patients' age split into five bins. Here, we calculate bias as the mean absolute deviation of the accuracy of each age bin from the overall accuracy. For the algorithm of \citet{monzannar_consistent_2020}, the average bias difference over all human expert predictions \(l\) between the proposed \textit{Embedding-SSL} approaches and the upper boundary decreases by 0.72 percentage points (pp) for gender and increases by 2.63 pp for age. Similarly, for the algorithm of \citet{raghu_algorithmic_2019}, we find a decrease of 1.53 pp for gender and an increase of 1.74 pp for age. Lastly, the algorithm of \citet{okati_differentiable_2021} has a 0.29 pp increase in bias for gender and a 0.8 pp increase in bias for age. We refer to the Appendix for additional visualizations for different numbers of human expert predictions \(l\) including an analysis for an additional radiologist. To summarize, we do not find evidence that the artificial expert predictions systematically amplify the bias of the evaluated learning to defer algorithms.

\section{Conclusion}
\label{subsec:conclusion}
In this work, we present a novel approach to reduce the number of human expert predictions required for the training of learning to defer algorithms. Our approach accurately learns the capabilities of an individual expert. On that basis, it generates artificial expert predictions that serve together with the available expert predictions as training data. The result of our empirical evaluation shows that even a small number of expert predictions per class are sufficient for generating artificial expert predictions that enable learning to defer algorithms to exceed the individual performance of both human and AI. Currently, the data instances used to learn the expert's capabilities are selected randomly. Future work could explore advanced strategies that determine representative instances for which an expert should provide predictions. This could further improve the effectiveness of our approach and potentially make it easier to onboard new experts to collaborate with learning to defer algorithms.

\section{Ethics Statement}

We hope that the proposed approach will contribute to the application of learning to defer algorithms in practice by reducing the required number of human expert predictions in addition to ground truth labels as a considerable implementation barrier. However, generating artificial expert predictions that are used together with a small number of human expert predictions to train learning to defer algorithms can entail the risk of potentially resulting in an overall system that is discriminatory against certain protected groups. For this reason, we want to raise awareness of the importance of monitoring fairness in the application settings in which the learning to defer algorithms are employed.

\section{Acknowledgments}

This work was supported by the Federal Ministry for Economic Affairs and Climate Action of Germany in the project Smart Design and Construction (project number 01MK20016F).

\bibliography{aaai23}

\end{document}